\documentclass[runningheads]{llncs}
\usepackage[T1]{fontenc}
\usepackage{booktabs}

\usepackage{graphicx}
\usepackage{amsmath,amsfonts,bm}

\def\eqref#1{equation~\ref{#1}}

\def\1{\bm{1}}

\def\vs{{\bm{s}}}

\def\vw{{\bm{w}}}

\def\mA{{\bm{A}}}

\def\mF{{\bm{F}}}

\def\mI{{\bm{I}}}

\def\mK{{\bm{K}}}

\def\mM{{\bm{M}}}

\def\mQ{{\bm{Q}}}

\def\mV{{\bm{V}}}
\def\mW{{\bm{W}}}

\DeclareMathAlphabet{\mathsfit}{\encodingdefault}{\sfdefault}{m}{sl}
\SetMathAlphabet{\mathsfit}{bold}{\encodingdefault}{\sfdefault}{bx}{n}

\def\sQ{{\mathbb{Q}}}

\def\sS{{\mathbb{S}}}

\usepackage{multirow} 
\usepackage{subcaption}
\usepackage{pifont}

\usepackage{colortbl}
\definecolor{gray}{RGB}{230,230,230}
\definecolor{lightyellow}{RGB}{255, 249, 192}
\definecolor{lightpurple}{RGB}{248, 231, 242}
\definecolor{timecolor}{RGB}{0, 115, 140}

\begin{document}
\title{Visual Memory and Textual Alignment for Cross-Modal Few-Shot Segmentation}
%
%
\titlerunning{ViTA-SAM}
%
\author{Tianqi Wei\inst{1} \and
Zhi Chen\inst{2} \and
Xin Yu\inst{3} \and
Scott Chapman\inst{1} \and
Zi Huang\inst{1}
}
\authorrunning{T. Wei et al.}
%
\institute{The University of Queensland, Brisbane, Australia \\
\email{\{tianqi.wei,scott.chapman,helen.huang\}@uq.edu.au}\\
 \and
University of Southern Queensland, Toowoomba, Australia\\
\email{Zhi.Chen@unisq.edu.au}
\and Adelaide University, Adelaide, Australia\\
\email{xin.yu@adelaide.edu.au}
}

\maketitle              
\begin{abstract}
Few-shot segmentation (FSS) aims to segment target objects with limited annotated examples. Recent studies have begun to explore large foundation models to improve generalization for this task. In particular, SAM~3 provides strong visual representations and supports text-guided segmentation. However, its performance remains limited in specialized domains, where semantic prompts and visual representations are often poorly aligned. To address this challenge, we propose ViTA-SAM, a framework built upon SAM~3 that integrates visual memory and textual features through cross-modal alignment. Specifically, we construct a visual memory module that encodes support features together with their corresponding masks, providing support-aware visual guidance for query prediction. In parallel, we leverage the text encoder of SAM~3 to extract semantic features as complementary priors. To bridge the gap between semantic and visual representations, we introduce a cross-modal alignment module that conditions textual features on memory-enhanced visual representations, yielding more consistent and visually grounded predictions. In addition, we propose a pseudo-label refinement strategy that updates the memory with predicted query masks, enabling query-adaptive representation learning. Extensive experiments on multiple benchmarks show that ViTA-SAM achieves the best overall performance among the compared methods, outperforming the strongest competing method by over 5\% in average mIoU.

\end{abstract}
\section{Introduction}
Semantic segmentation is a core task in computer vision and multimedia applications, with broad applications in medical imaging~\cite{ShiSS0YY24}, autonomous system~\cite{CaiSY24,YueLZHLNDZJCJ24}, and visual content understanding~\cite{SunZ024,WuLLYZZ24}. Recent advances in deep learning, together with the availability of large-scale annotated datasets, have led to substantial progress in segmentation performance \cite{cordts2016cityscapes,zhou2017ade20k}. Despite these successes, pixel-wise annotation is still highly labor-intensive and costly, making large-scale dataset construction difficult in practice.
Few-shot segmentation (FSS) \cite{siam2019amp,snell2017prototypical,wang2019panet,zhang2021selfguided,zhang2019canet} has therefore attracted increasing attention as a promising alternative. Its goal is to segment target objects from only a few annotated examples. 

\begin{figure*}[t]
    \centering
    \includegraphics[width=\textwidth]{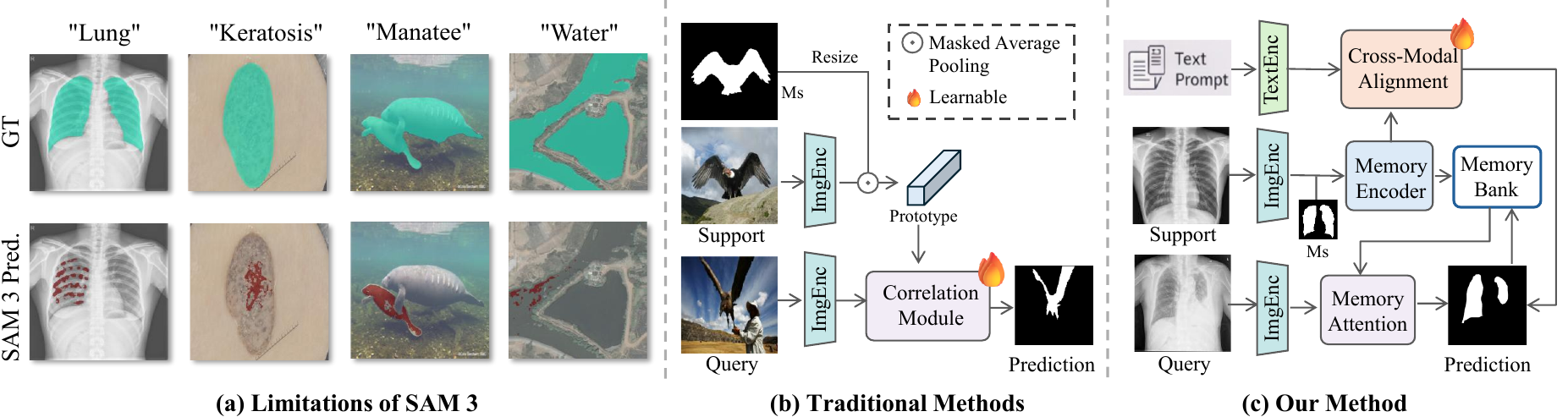}
    \caption{(a) Under text prompts, predictions show large deviations from the ground truth, revealing severe semantic–visual misalignment under domain shifts. (b) Traditional FSS methods rely on prototype matching between support and query features. (c) Our method introduces textual representations and performs cross-modal alignment with visual features to provide complementary semantic cues.}
    \label{fig:intro}
\end{figure*}

Conventional FSS methods mainly focus on constructing representative prototypes (Figure~\ref{fig:intro}b).
The pioneering work PANet~\cite{wang2019panet} learns class-specific prototypes from support images and segments the query image via pixel-wise matching in the feature space.
CANet~\cite{zhang2019canet} performs multi-level feature comparison between query features and global support representations.
PFENet~\cite{tian2020pfenet} leverages support-guided prior masks to encode coarse object locations and uses them to alleviate spatial inconsistency between support and query targets.
Despite their effectiveness, the visual matching functions learned by these methods are often biased toward the training distribution, limiting their generalization to diverse scenarios. Subsequent methods improve robustness to such distribution variations.
For example, PATNet~\cite{patnet} introduces a prototype alignment module to reduce feature discrepancies.
DMTNet~\cite{dmtnet} addresses distribution variations via a dynamic multi-scale mechanism that adaptively weights features based on input characteristics.
ABCDFSS~\cite{ABCDFSS} performs task-specific feature adaptation at test time, thereby better aligning representations with the current input.

Meanwhile, recent advances in large-scale foundation models, particularly the Segment Anything Model (SAM) family, have opened new opportunities for generalizable segmentation. The original SAM~\cite{kirillov2023sam} introduces promptable zero-shot segmentation using visual prompts such as points and bounding boxes. SAM~2~\cite{ravi2024sam2} extends this capability to video segmentation by incorporating a temporal memory mechanism for consistent mask propagation. More recently, SAM~3~\cite{carion2025sam3} integrates the memory mechanism from SAM~2 and improves vision--language alignment, enabling the use of both visual concepts and textual prompts. However, SAM models are primarily pretrained on natural images, which limits their generalization to specialized domains. As illustrated in Figure~\ref{fig:intro}a, SAM~3 struggles with specialized imagery (\textit{e.g.,} medical or satellite images) and domain-specific concepts (\textit{e.g.,} ``keratosis''), suggesting a misalignment between semantic and visual representations that degrades segmentation performance.

Based on these observations, we propose ViTA-SAM, a novel framework built on SAM~3 that aligns visual memory and textual prompts through cross-modal interaction and enables robust knowledge transfer (Figure~\ref{fig:intro}c). Specifically, we leverage the visual memory module to encode support features and their corresponding masks into a memory bank, producing object-aware representations. During inference, the memory is fused with query features to propagate structural and boundary information from support to query.
In parallel, we incorporate textual information using the SAM~3 text encoder to extract task-specific semantic features as complementary priors. To bridge the misalignment between semantic and visual representations under domain shifts, we align textual features with memory-enhanced visual features via cross-attention, yielding visually grounded semantic features.
Furthermore, we introduce a pseudo-label refinement strategy that uses predicted query masks to update the memory with query features, constructing a query-adaptive memory for progressively refined predictions. During inference, the memory-enhanced visual features and aligned textual features are fused using prediction-confidence-based weights.
Overall, the main contributions of this work are as follows:
\begin{itemize}
    \item We propose ViTA-SAM, a novel framework for few-shot segmentation. It integrates visual memory and textual features through cross-modal alignment, enabling effective knowledge transfer under domain shifts.
    \item We introduce a cross-modal alignment mechanism that aligns textual and visual representations via cross-attention, producing visually grounded semantic features and improving support-query alignment. A pseudo-label refinement strategy is also used to obtain query-adaptive representations.
    \item Extensive experiments on four widely used benchmarks demonstrate the effectiveness of our method, which generally outperforms existing approaches.
\end{itemize}

\section{Related Work}
\textbf{Few-shot Semantic Segmentation.}
Few-shot semantic segmentation (FSS) aims to segment query images using only a few annotated support samples. Existing methods can be broadly categorized into prototype-based and affinity-based methods. Prototype-based approaches \cite{dong2018few,li2021adaptive,lang2022learning,liu2022intermediate,snell2017prototypical,tian2020pfenet,wang2019panet} construct category prototypes from support images and segment queries by measuring feature similarity. However, prototype construction typically relies on masked average pooling, which may discard fine-grained spatial structures and limit representation capacity. Affinity-based methods \cite{lu2021simplerisbetter,min2021hypercorrelation,peng2023hierarchical,zhang2021few} model dense correspondences between query and support images, better preserving contextual and structural information. Despite their strong performance, traditional FSS approaches generally assume similar training and testing distributions, and often degrade under significant domain shifts.

\vspace{5pt}

\noindent\textbf{Cross-domain FSS.}
Cross-domain few-shot semantic segmentation (CD-FSS) considers a more challenging setting with significant domain shifts between source and target domains. Existing methods mainly address this problem by learning domain-invariant representations or adapting models to target-domain data. PATNet \cite{patnet}, DMTNet \cite{dmtnet}, and IFA \cite{IFA} model support-query relations to reduce domain-specific variations, while ABCDFSS \cite{ABCDFSS} and DR-Adapter~\cite{DR-Adapter} adapt pre-trained features or rectify domain styles at test time. APSeg \cite{apseg} further constructs query prototypes through cycle consistency and maps features into a stable domain-agnostic space. More recently, TVGTANet~\cite{liu2025TVGT} incorporates vision-language modeling by leveraging semantic information from prompts to guide CD-FSS.

\vspace{5pt}

\noindent\textbf{Foundation Models for Segmentation.}
Large-scale vision and vision-language foundation models, such as CLIP \cite{clip}, DINO \cite{oquab2023dinov2}, and LLaVA \cite{liu2024llava}, have demonstrated strong zero-shot and few-shot generalization across downstream tasks. The Segment Anything Model (SAM) family has emerged as a representative foundation model for promptable segmentation. SAM~\cite{kirillov2023sam} enables prompt-based zero-shot segmentation using visual cues, while SAM~2~\cite{ravi2024sam2} extends this capability to video through a temporal memory mechanism. More recently, SAM~3~\cite{carion2025sam3} incorporates vision-language alignment to support open-vocabulary segmentation. Despite their strong zero-shot capabilities, these models remain constrained by significant domain shifts in specialized scenarios. 
To bridge this gap, we propose ViTA-SAM, which combines the visual representations and textual guidance of SAM~3 with annotated support images for few-shot segmentation.

\section{Methodology}

\subsection{Task Formulation}
Following the meta-learning paradigm~\cite{patnet}, the FSS task is formulated as learning from a series of episodes. Each episode consists of a query set $\sQ = \{ (\mI^q, \mM^q) \}$ and a support set $\sS = \{ (\mI^s_k, \mM^s_k) \}_{k=1}^K$, where both sets belong to the same category.
We denote $\mI \in \mathbb{R}^{H \times W \times 3}$ as the input image and use $\mM \in \{0, 1\}^{H \times W}$ to represent the corresponding binary ground-truth mask. $K$ represents the shot number, indicating the number of annotated support samples available to guide the segmentation of the query image $\mI^q$. 
The objective is to acquire a segmentation model $f_{\theta}$ that predicts the query mask $\hat{\mM}^q = f_{\theta}(\mI^q, \sS)$.
During the inference process, the model is evaluated on episodes sampled from the test set $\mathcal{D}_{test}$.

\begin{figure*}[t]
    \centering
    \includegraphics[width=\textwidth]{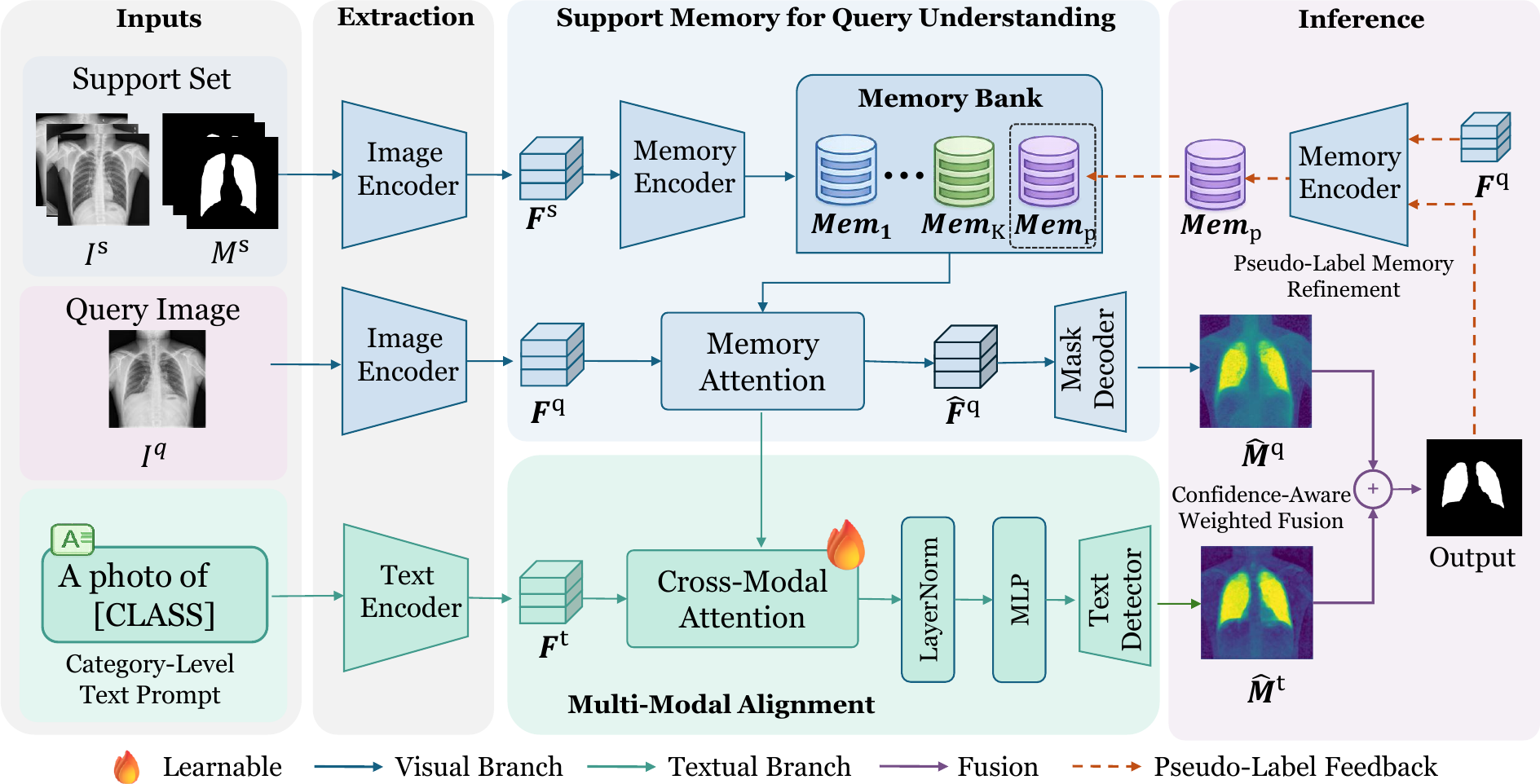}
    \caption{Overview of ViTA-SAM. The method integrates visual memory and textual alignment to capture both appearance and semantic cues. 
    }
    \label{fig:overview}
\end{figure*}

\subsection{ViTA-SAM}
The multi-modal alignment capability of SAM~3 forms the representational basis of our ViTA-SAM by projecting different modalities into a shared latent space. As illustrated in Figure~\ref{fig:overview}, given a support image, a query image $\mI^q$, and a category-level text prompt $T$, the image encoder $\bm{IE}$ first extracts the support feature $\mF^{s}$ and the query feature $\mF^{q}$, while the text encoder $\bm{TE}$ produces the textual feature $\mF^{t}$. The support feature is then encoded into a memory bank $\mathcal{B}$, from which the query feature retrieves relevant information to obtain an enhanced representation for segmentation.
In parallel, the textual feature is conditioned on the memory-enhanced visual representation through a lightweight cross-attention module. To facilitate cross-modal alignment, this module is trained using pseudo-labels derived from the output of the visual memory branch. Furthermore, to improve prediction consistency, we introduce a pseudo-label refinement step that updates the memory with query-specific information derived from the current prediction.

\subsection{Support Memory for Query Understanding}
The visual memory module aims to extract and preserve appearance-level cues from annotated support images and transfer these cues to the query representation. Unlike traditional prototypical approaches that compress support information into a single vector, our design maintains spatially-aware feature representations to provide fine-grained guidance during query segmentation.

\vspace{5pt}
\noindent \textbf{Memory Construction.} 
Given $K$-shot support images  $\{\mI^s_k\}_{k=1}^K$, we first use the SAM~3 image encoder $\bm{IE}$ to obtain the visual features $\mF_k^s = \bm{IE}(\mI_k^s) \in \mathbb{R}^{h \times w \times C}$. Then, the memory encoder $\bm{ME}$ encodes the image features with their corresponding masks as memories:
\begin{equation}
    \bm{Mem}_k = \bm{ME}(\mF^s_k ,\mM^s_k), \quad k \in \{1, \dots, K\}.
\end{equation}
The visual memory bank $\mathcal{B}$ is then defined as the collection of these mask-conditioned representations:
\begin{equation}
    \mathcal{B} =  Concat( {\bm{Mem}}_1, {\bm{Mem}}_2, \dots, {\bm{Mem}}_K ).
\end{equation}

\vspace{5pt}
\noindent \textbf{Query Interaction.} 
The query interaction enables the query to retrieve spatially relevant support cues from the memory and incorporate appearance priors into its representation.
For a query image $\mI^q$, the image encoder extracts the features $\mF^q \in \mathbb{R}^{hw \times C}$. The core of our guidance mechanism lies in the fusion of $\mF^q$ with the structured memory $\mathcal{B}$ in the memory attention $\bm{MA}(\cdot)$ mechanism which is implemented as the cross-attention as:
\begin{equation}
\mQ^{ma} = \mF^q \mW_q^{\top}, 
\mK^{ma} = \mathcal{B}\mW_k^{\top}, 
\mV^{ma} = \mathcal{B}\mW_v^{\top},
\end{equation}
\begin{equation}
\mA^{ma} = \mathrm{softmax}\!\left(\mQ^{ma}\mK^{ma\top}/\sqrt{d}\right), 
\hat{\mF}^q = \bm{MA}(\mF^q, \mathcal{B}) = \mA^{ma}\mV^{ma}, \nonumber
\end{equation}
where $d$ is the attention dimension. In this way, the query tokens serve as queries, while the concatenated support memories in $\mathcal{B}$ act as the keys and values for feature aggregation. The resulting enhanced query representation $\hat{\mF}^q$ is then reshaped back to the spatial layout and fed into the mask decoder of SAM~3 to produce the segmentation mask logits $\hat{\mM}^q$.

\subsection{Cross-Modal Alignment}

\noindent \textbf{Textual Feature Extraction.} 
While the support memory captures appearance-level cues from annotated support images, semantic priors provide complementary category-level information. To incorporate such semantic guidance, we extract textual representations using the text encoder $\bm{TE}$ of SAM~3. Given a textual prompt $T$, the text encoder produces a semantic embedding:
\begin{equation}
\mF^t = \bm{TE}(T) \in \mathbb{R}^{L \times C},
\end{equation}
where $L$ denotes the number of text tokens. We keep the text encoder frozen to preserve its generalization ability. The resulting embedding encodes category-level semantics and serves as a complementary source of guidance for segmentation. Compared with the support memory, which mainly conveys instance-level visual cues, the textual feature provides more abstract semantic information and is generally more robust to appearance variations across domains.

\vspace{5pt}
\noindent \textbf{Grounding Text with Visual Memory.} 
In cross-modal few-shot segmentation, semantic priors often fail to align with visual patterns under significant domain shifts, leading to inconsistent predictions. To address this issue, we introduce a lightweight cross-attention module that aligns the textual features with the memory-enhanced visual representations. Let the textual feature be denoted by $\mF^t \in \mathbb{R}^{L \times C}$, and let the memory-enhanced query feature be denoted by $\hat{\mF}^q \in \mathbb{R}^{hw \times C}$. We construct the cross-modal alignment module $\bm{CM}(\cdot)$ in the cross-attention form:
\begin{equation}
\mQ^{ta} = \mF^t \tilde{\mW}_q^{\top}, 
\mK^{ta} = \hat{\mF}^q \tilde{\mW}k^{\top}, 
\mV^{ta} = \hat{\mF}^q \tilde{\mW}_v^{\top},
\end{equation}
\begin{equation}
\mA^{ta} = \mathrm{softmax}\!\left({\mQ^{ta}\mK^{ta\top}}/{\sqrt{d}}\right), ~
\hat{\mF}^t = \bm{CM}(\mF^t, \hat{\mF}^q) = \mA^{ta}\mV^{ta},
\end{equation}

where $d$ denotes the attention dimension. In this way, the textual feature serves as the query, while the memory-enhanced visual representation provides the keys and values. This operation conditions the semantic representation on support-aware visual cues, producing a visually grounded textual representation. The updated feature is then processed by a normalization layer and a feed-forward network, and finally fed into the frozen detector of SAM~3 to generate the mask logits for the textual branch, denoted by $\hat{\mM}^t$.

To encourage consistent cross-modal alignment, we supervise the textual branch using the output of the memory-based visual prediction. Specifically, the predicted mask logits $\hat{\mM}^q$ are binarized to obtain a pseudo mask $\mM^p$, which serves as supervision for the textual branch. The training objective is defined as
\begin{equation}
\mathcal{L}_{\mathrm{CE}} = \mathrm{BCE}(\hat{\mM}^t, \mM^p).
\end{equation}
This supervision encourages the textual representation to converge toward the memory-guided visual prediction, thereby reducing cross-modal discrepancies and improving robustness under severe domain shifts.

\subsection{Inference}
\noindent \textbf{Dual-Branch Prediction.} 
During inference, we combine predictions from the visual memory and textual branches using a confidence-aware fusion strategy. Let $\hat{\mM}^q$ and $\hat{\mM}^t$ denote their mask logits. The confidence of each branch is estimated by the average logit magnitude:
\begin{equation}
\vs_q = \operatorname{GAP}(|\hat{\mM}^q|), \qquad
\vs_t = \operatorname{GAP}(|\hat{\mM}^t|),
\end{equation}
where $\operatorname{GAP}(\cdot)$ denotes global average pooling. The normalized fusion weights are
\begin{equation}
\vw_q = \frac{\vs_q}{\vs_q + \vs_t}, \qquad
\vw_t = \frac{\vs_t}{\vs_q + \vs_t}.
\end{equation}
The fused prediction is computed as
\begin{equation}
\hat{\mM} = \vw_q \cdot \sigma(\hat{\mM}^q) + \vw_t \cdot \sigma(\hat{\mM}^t),
\end{equation}
where $\sigma(\cdot)$ denotes the Sigmoid function. The initial binary mask is obtained as
$\tilde{\mM} = \mathbb{I}(\hat{\mM} > 0.5)$.

\vspace{5pt}

\noindent \textbf{Pseudo-Label Memory Refinement.} 
We further use the initial prediction $\tilde{\mM}$ as a pseudo label to construct query-adaptive memory:
\begin{equation}
\bm{Mem}_{p} = \bm{ME}(\mF^q, \tilde{\mM}),
\end{equation}
where $\mathrm{ME}(\cdot)$ denotes the memory encoder of SAM~3. The support memory is then augmented with the pseudo-labeled query memory:
\begin{equation}
\mathcal{B}' = \operatorname{Concat}(
{\bm{Mem}}_1, {\bm{Mem}}_2, \dots, {\bm{Mem}}_K, \bm{Mem}_{p}).
\end{equation}
The updated memory bank $\mathcal{B}'$ is fed back into the visual memory branch to refine the query representation and prediction.

\section{Experiment}

\subsection{Datasets}
We evaluate our method on four widely used benchmarks covering diverse target domains: ISIC2018~\cite{codella2019skinisic}, Chest X-ray~\cite{candemir2013lung1}, DeepGlobe~\cite{demir2018deepglobe}, and FSS-1000~\cite{li2020fss}.
ISIC2018 contains 2,596 images from three skin lesion categories and presents challenges in fine-grained texture discrimination. Chest X-ray consists of 566 annotated lung X-ray images characterized by low contrast and structural noise. DeepGlobe provides high-resolution satellite imagery for land-cover segmentation. Following~\cite{patnet}, we exclude the `unknown' class and use the remaining six categories (5,666 images). FSS-1000 contains 1,000 natural object classes, and we evaluate on its official test set of 2,400 images from 240 classes.


\subsection{Implementation Details.} 
To preserve the foundational segmenting capabilities, we keep all SAM~3 components frozen and adopt a parameter-efficient adaptation strategy by optimizing only the proposed cross-modal alignment module.
We adopt the Adam optimizer with a learning rate of $1\times10^{-4}$ and supervise the model using binary cross-entropy loss.
To align with the original image input setting of SAM~3, images are resized to $1024\times1024$, with a batch size of 1.
For text input, we use template-based prompts constructed from category names
Following the standard evaluation protocol in prior work~\cite{wang2019panet}, we use mean Intersection over Union (mIoU) as the evaluation metric.

\subsection{Main Results}
\noindent \textbf{Comparison with Existing Methods.} 
The proposed method is comprehensively compared with state-of-the-art approaches for both traditional few-shot semantic segmentation and cross-domain few-shot semantic segmentation. The results are presented in Table~\ref{tab:main_results}.
Traditional FSS methods perform competitively on FSS-1000, which mainly contains natural images, but generally suffer substantial performance degradation on specialized target domains.
In contrast, ViTA-SAM achieves the best average performance, surpassing the previous best results by 6.13\% and 4.69\% in average mIoU under the 1-shot and 5-shot settings, respectively.
It achieves the best performance on Chest X-ray, ISIC, and FSS-1000, demonstrating strong generalization across diverse target domains.
Particularly, notable gains are observed on the medical datasets Chest X-ray and ISIC, demonstrating the effectiveness of our method under substantial domain shifts.
On the challenging Deepglobe benchmark, ViTA-SAM still achieves the best 1-shot result and remains highly competitive in the 5-shot setting.
Moreover, the larger overall improvement under the 1-shot setting suggests that ViTA-SAM is particularly effective when only limited support supervision is available.
Overall, these results demonstrate the robustness and generalization capability of our method across diverse domains and few-shot settings.

\begin{table*}[t]
\centering
\caption{Quantitative comparison of the proposed \textbf{ViTA-SAM} with existing few-shot and cross-domain few-shot segmentation methods in mIoU (\%).}
\label{tab:main_results}
\resizebox{\textwidth}{!}{%
\begin{tabular}{l|cc|cc|cc|cc|cc}
\toprule
\multirow{2}{*}{Methods} & \multicolumn{2}{c|}{Chest X-ray} & \multicolumn{2}{c|}{ISIC} & \multicolumn{2}{c|}{FSS-1000} & \multicolumn{2}{c|}{Deepglobe} & \multicolumn{2}{c}{Average} \\
\cline{2-3} \cline{4-5} \cline{6-7} \cline{8-9} \cline{10-11}
& 1-shot & 5-shot & 1-shot & 5-shot & 1-shot & 5-shot & 1-shot & 5-shot & 1-shot & 5-shot \\
\midrule
\rowcolor{lightyellow} \multicolumn{11}{c}{Traditional Few-shot Semantic Segmentation Methods} \\
\midrule
AMP \cite{siam2019amp} & 51.23 & 53.04 & 28.42 & 30.41 & 57.18 & 59.24 & 37.61 & 40.61 & 43.61 & 45.83 \\
PANet \cite{wang2019panet} & 57.75 & 69.31 & 25.29 & 33.99 & 69.15 & 71.68 & 36.55 & 45.43 & 47.19 & 55.10 \\
CANet \cite{zhang2019canet} & 28.35 & 28.62 & 26.12 & 29.22 & 70.67 & 72.03 & 20.37 & 23.03 & 36.38 & 38.23 \\ 
RPMMs \cite{yang2020rpmms} & 30.11 & 30.15 & 18.02 & 20.04 & 65.12 & 67.06 & 12.99 & 13.47 & 31.56 & 32.68 \\ 
PFENet \cite{tian2020pfenet} & 27.22 & 27.57 & 25.77 & 27.35 & 70.52 & 72.16 & 10.52 & 11.86 & 33.51 & 34.74 \\ 
HSNet \cite{min2021hypercorrelation} & 51.88 & 54.36 & 31.96 & 35.33 & 75.50 & 77.53 & 29.65 & 35.45 & 47.25 & 50.67 \\ 
PerSAM \cite{zhang2023persam} & 29.95 & 30.05 & 23.27 & 25.33 & 60.92 & 66.53 & 36.08 & 40.65 & 37.56 & 40.64 \\
\midrule
\rowcolor{lightyellow} \multicolumn{11}{c}{Cross-domain Few-shot Semantic Segmentation Methods} \\
\midrule
PATNet \cite{patnet} & 66.61 & 70.20 & 41.16 & 53.58 & 78.59 & 81.23 & 37.89 & 42.97 & 56.06 & 61.99 \\
DMTNet \cite{dmtnet} & 73.74 & 77.30 & 43.55 & 52.30 & 81.52 & 83.28 & 40.14 & 51.17 & 59.74 & 66.01 \\
APSeg \cite{apseg} & 84.10 & 84.50 & 45.43 & 53.98 & 79.71 & 81.90 & 35.94 & 39.98 & 61.30 & 65.09 \\
DR-Adapter \cite{DR-Adapter} & 82.35 & 82.35 & 40.77 & 48.87 & 79.05 & 80.40 & 41.29 & 50.12 & 60.86 & 65.42 \\
ABCDFSS \cite{ABCDFSS} & 79.80 & 81.40 & 45.70 & 53.30 & 74.60 & 76.20 & 42.60 & 49.00 & 60.68 & 64.98 \\
ISA \cite{Fan_2025_ICCV_isa} & 83.40 & 86.30 & 37.20 & 56.10 & 78.80 & 86.00 & 44.30 & \textbf{52.10} & 60.93 & 70.13 \\
TVGTANet \cite{liu2025TVGT} & 84.58 & 87.27 & 47.21 & 58.75 & 78.32 & 81.44 & 42.04 & 50.67 & 63.04 & 69.53 \\
DFN \cite{tong2025adapterDFN} & 83.18 & 87.14 & 50.36 & 58.53 & 82.97 & 85.72 & 39.45 & 47.67 & 63.99 & 69.77 \\
\midrule
\rowcolor{lightpurple} \textbf{ViTA-SAM (Ours)} & \textbf{90.53} & \textbf{91.97} & \textbf{57.80} & \textbf{66.42} & \textbf{87.71} & \textbf{88.83} & \textbf{44.42} & {52.04} & \textbf{70.12} & \textbf{74.82} \\
\bottomrule
\end{tabular}
}
\end{table*}

\begin{figure*}[t]
    \centering
    \includegraphics[width=\textwidth]{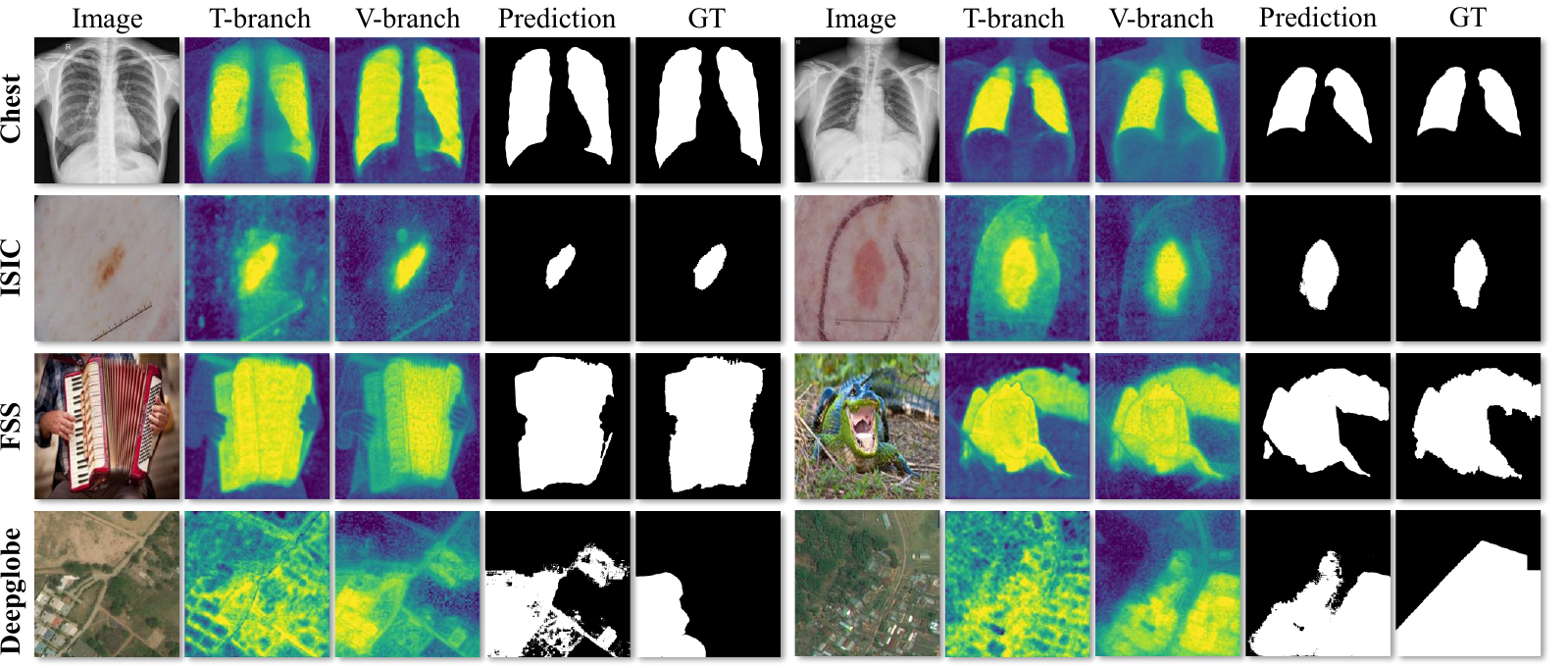}
    \caption{Visualization of output logit maps of the textual and visual modules.
    From left to right: input images, logits from the aligned text module (T-branch), logits from the visual memory module (V-branch), final prediction, and GT masks.}
    \label{fig:vis_logits}
\end{figure*}

\vspace{5pt}
\noindent \textbf{Qualitative Analysis of Different Modules.} 
\noindent Figure~\ref{fig:vis_logits} presents qualitative comparisons of output logit maps from different modules.
When target regions are relatively regular and semantically clear, such as in Chest X-ray and FSS, both branches respond effectively, while the textual branch produces smoother and more concentrated activations.
On ISIC, both branches can localize the lesion, but the visual branch yields a more compact and cleaner response.
On Deepglobe, the textual branch produces scattered activations over irrelevant regions, whereas the visual branch provides more stable localization despite residual background noise.
These observations highlight the complementary roles of the two branches: the textual branch offers stronger semantic emphasis, while the visual branch provides more reliable spatial grounding in complex scenarios.

\subsection{Ablation Study}

\noindent \textbf{Effectiveness of Specific Modules.} 
Table~\ref{tab:ablation_branch} analyzes the individual and combined contributions of the core components in {ViTA-SAM}. Results are reported under the 1-shot setting.
The visual memory module (Visual) alone achieves strong performance by effectively transferring appearance-level cues from support to query via the memory attention mechanism of SAM~3.
The textual branch, which produces memory-aligned semantic features through cross-modal alignment, also achieves competitive results, indicating that semantic guidance provides complementary information beyond pure visual matching.
Distinct behaviors are observed across datasets. The textual branch performs better on the Chest X-ray dataset, while the visual memory module shows stronger performance on other benchmarks, highlighting its advantage in modeling complex visual patterns and fine-grained structures.
Combining both components consistently improves performance across all benchmarks, confirming their complementary roles in aligning appearance and semantic representations.
Further incorporating pseudo-label refinement (Refinement) leads to additional gains, demonstrating the effectiveness of leveraging semantically confident predictions to enhance the visual memory.


\begin{table}[t]
\centering
\caption{Ablation study of the proposed modules. 
Visual: visual memory branch; Textual: aligned textual branch; Refine: pseudo-label refinement.}
\label{tab:ablation_branch}
\begin{tabular}{ccc | cccc}
\toprule
\textbf{Visual} & \textbf{Textual} & \textbf{Refine} & \textbf{CXR} & \textbf{ISIC} & \textbf{FSS} & \textbf{DGL} \\
\midrule
\ding{51} & \ding{55} & \ding{55} & 87.28 & 53.36 & 85.37 & 43.10 \\
\ding{55} & \ding{51} & \ding{55} & 88.64 & 46.95 & 83.90 & 36.49 \\
\ding{51} & \ding{51} & \ding{55} & 90.29 & 56.98 & 87.60 & 43.90 \\
\ding{51} & \ding{51} & \ding{51} & \textbf{90.53} & \textbf{57.80} & \textbf{87.71} & \textbf{44.42} \\
\bottomrule
\end{tabular}
\end{table}

\vspace{5pt}
\noindent \textbf{Multi-step Pseudo-label Refinement.} 
To examine the effectiveness of pseudo-label refinement, we conduct a multi-step study.
Starting from the initial prediction $\tilde{\mM}$, the predicted mask is used as a pseudo-label and re-encoded into the memory to construct a query-adaptive memory.
The updated memory is then combined with query features to produce a refined prediction.
This process can be repeated iteratively, resulting in $\tilde{\mM}_n$ after $n$ steps and progressively refining the query representations.
We evaluate the impact of different refinement steps on performance, as reported in Table~\ref{tab:iteration}.
A single refinement step ($n=1$) consistently improves performance across all benchmarks.
This gain comes from introducing query-derived pseudo-labels, which augment the support-only memory with query-specific cues and improve alignment with the target distribution.
Increasing the number of refinement steps does not yield consistent gains.
While slight improvements are observed on Chest X-ray, performance saturates or degrades on other datasets.
This behavior is mainly due to error accumulation in pseudo-labels: repeatedly encoding self-predicted masks introduces noisy signals into the memory, degrading feature alignment.
This effect is less pronounced on Chest X-ray, where pseudo-labels are more reliable.
In contrast, datasets with lower pseudo-label quality are more susceptible to noise accumulation, resulting in performance degradation with multiple refinement steps.
Therefore, we adopt a single refinement step ($n=1$) as the default setting.

\begin{table}[t]
\centering
\caption{Effect of iterative pseudo-label refinement.}
\label{tab:iteration}

\begin{tabular}{c | cccc}
\toprule
\textbf{Iteration} & \textbf{CXR} & \textbf{ISIC} & \textbf{FSS} & \textbf{DGL} \\
\midrule
0 & 90.29 & 56.98 & 87.60 & 43.90\\
1  & 90.53 & \textbf{57.80} & \textbf{87.71} & \textbf{44.42} \\
2 & 90.59 & 57.79 & 87.52 & 44.11 \\
3 & \textbf{90.61} & 57.20 & 87.13 & 44.09\\
\bottomrule
\end{tabular}

\end{table}

\subsection{Discussion}

\noindent \textbf{Effectiveness of Cross-modal Alignment.}
SAM~3 supports text-driven segmentation by taking textual prompts as input and predicting corresponding regions. However, under significant domain shifts, such as medical and satellite imagery, textual semantics can be poorly aligned with target-domain visual patterns.
We therefore compare the raw outputs of the SAM~3 text branch with those obtained after introducing the proposed cross-modal alignment mechanism, which incorporates memory-enhanced visual representations. As shown in Table~\ref{tab:raw}, the original SAM~3 achieves moderate performance on FSS, which contains relatively common object categories, but performs poorly on the other datasets. In particular, the results on ISIC and Chest X-ray are close to zero, indicating that textual prompts alone provide insufficient guidance under substantial domain shifts.
After introducing the proposed alignment strategy, performance improves consistently across all benchmarks, demonstrating the importance of grounding textual representations with memory-enhanced visual features for reliable cross-domain segmentation.

\begin{table}[t]
\centering
\caption{Comparison of the SAM~3 textual branch performance before and after the proposed cross-modal alignment. }
\label{tab:raw}

\begin{tabular}{c | cccc}
\toprule
\textbf{Method} & \textbf{CXR} & \textbf{ISIC} & \textbf{FSS} & \textbf{DGL} \\
\midrule
SAM~3 (Text)  & 0.66 & 0.65 & 51.04 & 6.38 \\
SAM~3 (Text + Alignment)  & \textbf{88.64} & \textbf{46.95} & \textbf{83.90} & \textbf{36.49} \\
\bottomrule

\end{tabular}

\end{table}

\begin{figure}[t]
    \centering
    \includegraphics[width=0.6\textwidth]{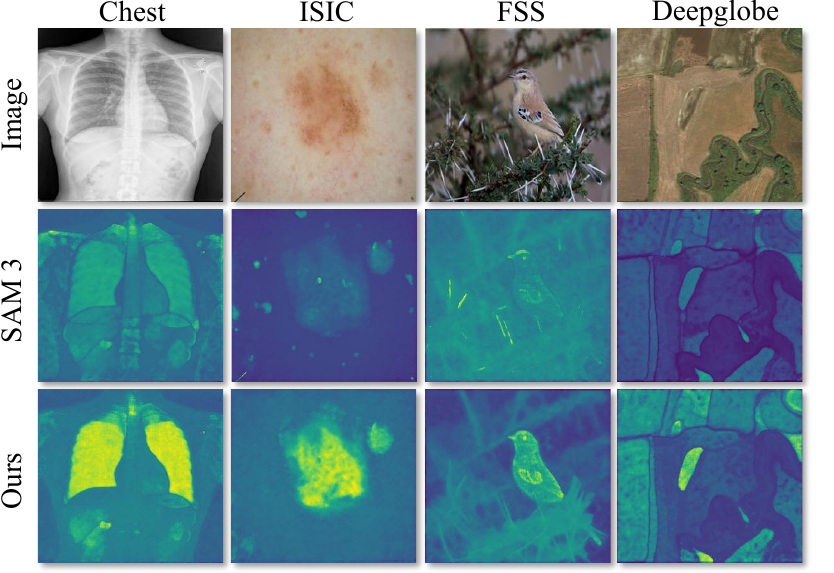}
    \caption{Comparison of response maps from raw SAM~3 and ViTA-SAM.}
    \label{fig:bf_af}
\end{figure}

\vspace{5pt}
\noindent \textbf{Visualization of Text–Visual Alignment.}
To analyze cross-modal alignment, we visualize response maps across benchmarks in Figure~\ref{fig:bf_af}. The first row displays input images, while the subsequent rows compare the text-visual response maps of raw SAM~3 and our aligned model.
Without cross-modal alignment, SAM~3 produces weak and spatially diffuse activations that fail to localize target regions, particularly under large domain shifts. In contrast, our method yields more concentrated and consistent activations with clearer foreground–background separation.
These results indicate that aligning textual semantics with memory-enhanced visual representations produces more reliable responses and improves spatial localization.

\section{Conclusion}

In this work, we present ViTA-SAM, a cross-modal few-shot semantic segmentation framework built upon the SAM~3 foundation model. Our approach integrates visual memory with textual features, where the visual memory provides support-aware structural guidance and the textual features introduce complementary semantic priors. We introduce a lightweight cross-modal interaction mechanism that conditions textual features on memory-enhanced visual representations, producing more consistent and visually grounded predictions. In addition, we introduce a pseudo-label refinement strategy that updates the memory using predicted query masks, enabling query-adaptive representation learning. Extensive experiments on multiple benchmarks show that ViTA-SAM consistently improves over existing methods.

\subsubsection{\ackname} 

This work was partially supported by the Grains Research and Development Corporation (UOQ2301-010OPX, USQ2605-002BGX) and by The University of Queensland (DVCR2201A).

\clearpage

%
%
%
\bibliographystyle{splncs04}
\bibliography{sample-base}

@String{Computer = "{IEEE} Computer" }

@String{Springer = "Springer-Verlag" }

@inproceedings{dong2018few,
  title={Few-shot semantic segmentation with prototype learning.},
  author={Dong, Nanqing and Xing, Eric P},
  booktitle={BMVC},
  volume={3},
  number={4},
  pages={4},
  year={2018}
}

@inproceedings{lang2022learning,
  title={Learning what not to segment: A new perspective on few-shot segmentation},
  author={Lang, Chunbo and Cheng, Gong and Tu, Binfei and Han, Junwei},
  booktitle={Proceedings of the IEEE/CVF conference on computer vision and pattern recognition},
  pages={8057--8067},
  year={2022}
}

@inproceedings{li2021adaptive,
  title={Adaptive prototype learning and allocation for few-shot segmentation},
  author={Li, Gen and Jampani, Varun and Sevilla-Lara, Laura and Sun, Deqing and Kim, Jonghyun and Kim, Joongkyu},
  booktitle={Proceedings of the IEEE/CVF conference on computer vision and pattern recognition},
  pages={8334--8343},
  year={2021}
}

@article{liu2022intermediate,
  title={Intermediate prototype mining transformer for few-shot semantic segmentation},
  author={Liu, Yuanwei and Liu, Nian and Yao, Xiwen and Han, Junwei},
  journal={Advances in Neural Information Processing Systems},
  volume={35},
  pages={38020--38031},
  year={2022}
}

@article{snell2017prototypical,
  title={Prototypical networks for few-shot learning},
  author={Snell, Jake and Swersky, Kevin and Zemel, Richard},
  journal={Advances in neural information processing systems},
  volume={30},
  year={2017}
}

@inproceedings{wang2019panet,
  title={Panet: Few-shot image semantic segmentation with prototype alignment},
  author={Wang, Kaixin and Liew, Jun Hao and Zou, Yingtian and Zhou, Daquan and Feng, Jiashi},
  booktitle={proceedings of the IEEE/CVF international conference on computer vision},
  pages={9197--9206},
  year={2019}
}

@inproceedings{lu2021simplerisbetter,
  title={Simpler is better: Few-shot semantic segmentation with classifier weight transformer},
  author={Lu, Zhihe and He, Sen and Zhu, Xiatian and Zhang, Li and Song, Yi-Zhe and Xiang, Tao},
  booktitle={Proceedings of the IEEE/CVF International Conference on Computer Vision},
  pages={8741--8750},
  year={2021}
}

@inproceedings{min2021hypercorrelation,
  title={Hypercorrelation squeeze for few-shot segmentation},
  author={Min, Juhong and Kang, Dahyun and Cho, Minsu},
  booktitle={Proceedings of the IEEE/CVF international conference on computer vision},
  pages={6941--6952},
  year={2021}
}

@inproceedings{peng2023hierarchical,
  title={Hierarchical dense correlation distillation for few-shot segmentation},
  author={Peng, Bohao and Tian, Zhuotao and Wu, Xiaoyang and Wang, Chengyao and Liu, Shu and Su, Jingyong and Jia, Jiaya},
  booktitle={Proceedings of the IEEE/CVF Conference on Computer Vision and Pattern Recognition},
  pages={23641--23651},
  year={2023}
}

@article{zhang2021few,
  title={Few-shot segmentation via cycle-consistent transformer},
  author={Zhang, Gengwei and Kang, Guoliang and Yang, Yi and Wei, Yunchao},
  journal={Advances in Neural Information Processing Systems},
  volume={34},
  pages={21984--21996},
  year={2021}
}

@inproceedings{clip,
  title={Learning transferable visual models from natural language supervision},
  author={Radford, Alec and Kim, Jong Wook and Hallacy, Chris and Ramesh, Aditya and Goh, Gabriel and Agarwal, Sandhini and Sastry, Girish and Askell, Amanda and Mishkin, Pamela and Clark, Jack and others},
  booktitle={International conference on machine learning},
  pages={8748--8763},
  year={2021},
  organization={PMLR}
}

@article{oquab2023dinov2,
  title={Dinov2: Learning robust visual features without supervision},
  author={Oquab, Maxime and Darcet, Timoth{\'e}e and Moutakanni, Th{\'e}o and Vo, Huy and Szafraniec, Marc and Khalidov, Vasil and Fernandez, Pierre and Haziza, Daniel and Massa, Francisco and El-Nouby, Alaaeldin and others},
  journal={arXiv preprint arXiv:2304.07193},
  year={2023}
}

@article{liu2024llava,
  title={Visual instruction tuning},
  author={Liu, Haotian and Li, Chunyuan and Wu, Qingyang and Lee, Yong Jae},
  journal={Advances in neural information processing systems},
  volume={36},
  year={2024}
}

@inproceedings{kirillov2023sam,
  title={Segment anything},
  author={Kirillov, Alexander and Mintun, Eric and Ravi, Nikhila and Mao, Hanzi and Rolland, Chloe and Gustafson, Laura and Xiao, Tete and Whitehead, Spencer and Berg, Alexander C and Lo, Wan-Yen and others},
  booktitle={Proceedings of the IEEE/CVF International Conference on Computer Vision},
  pages={4015--4026},
  year={2023}
}

@article{ravi2024sam2,
  title={Sam 2: Segment anything in images and videos},
  author={Ravi, Nikhila and Gabeur, Valentin and Hu, Yuan-Ting and Hu, Ronghang and Ryali, Chaitanya and Ma, Tengyu and Khedr, Haitham and R{\"a}dle, Roman and Rolland, Chloe and Gustafson, Laura and others},
  journal={arXiv preprint arXiv:2408.00714},
  year={2024}
}

@inproceedings{apseg,
  title={APSeg: Auto-Prompt Network for Cross-Domain Few-Shot Semantic Segmentation},
  author={He, Weizhao and Zhang, Yang and Zhuo, Wei and Shen, Linlin and Yang, Jiaqi and Deng, Songhe and Sun, Liang},
  booktitle={Proceedings of the IEEE/CVF Conference on Computer Vision and Pattern Recognition},
  pages={23762--23772},
  year={2024}
}

@inproceedings{ifa,
  title={Cross-Domain Few-Shot Segmentation via Iterative Support-Query Correspondence Mining},
  author={Nie, Jiahao and Xing, Yun and Zhang, Gongjie and Yan, Pei and Xiao, Aoran and Tan, Yap-Peng and Kot, Alex C and Lu, Shijian},
  booktitle={Proceedings of the IEEE/CVF Conference on Computer Vision and Pattern Recognition},
  pages={3380--3390},
  year={2024}
}

@inproceedings{abcdfss,
  title={Adapt Before Comparison: A New Perspective on Cross-Domain Few-Shot Segmentation},
  author={Herzog, Jonas},
  booktitle={Proceedings of the IEEE/CVF Conference on Computer Vision and Pattern Recognition},
  pages={23605--23615},
  year={2024}
}

@inproceedings{DR-Adapter,
  title={Domain-Rectifying Adapter for Cross-Domain Few-Shot Segmentation},
  author={Su, Jiapeng and Fan, Qi and Pei, Wenjie and Lu, Guangming and Chen, Fanglin},
  booktitle={Proceedings of the IEEE/CVF Conference on Computer Vision and Pattern Recognition},
  pages={24036--24045},
  year={2024}
}

@inproceedings{patnet,
  title={Cross-domain few-shot semantic segmentation},
  author={Lei, Shuo and Zhang, Xuchao and He, Jianfeng and Chen, Fanglan and Du, Bowen and Lu, Chang-Tien},
  booktitle={European Conference on Computer Vision},
  pages={73--90},
  year={2022},
  organization={Springer}
}

@inproceedings{dmtnet,
  title     = {Cross-Domain Few-Shot Semantic Segmentation via Doubly Matching Transformation},
  author    = {Chen, Jiayi and Quan, Rong and Qin, Jie},
  booktitle = {Proceedings of the Thirty-Third International Joint Conference on
               Artificial Intelligence, {IJCAI-24}},
  pages     = {641--649},
  year      = {2024},
}

@inproceedings{zhang2021selfguided,
  title={Self-guided and cross-guided learning for few-shot segmentation},
  author={Zhang, Bingfeng and Xiao, Jimin and Qin, Terry},
  booktitle={Proceedings of the IEEE/CVF conference on computer vision and pattern recognition},
  pages={8312--8321},
  year={2021}
}

@inproceedings{cordts2016cityscapes,
  title={The cityscapes dataset for semantic urban scene understanding},
  author={Cordts, Marius and Omran, Mohamed and Ramos, Sebastian and Rehfeld, Timo and Enzweiler, Markus and Benenson, Rodrigo and Franke, Uwe and Roth, Stefan and Schiele, Bernt},
  booktitle={Proceedings of the IEEE conference on computer vision and pattern recognition},
  pages={3213--3223},
  year={2016}
}

@inproceedings{zhou2017ade20k,
  title={Scene parsing through ade20k dataset},
  author={Zhou, Bolei and Zhao, Hang and Puig, Xavier and Fidler, Sanja and Barriuso, Adela and Torralba, Antonio},
  booktitle={Proceedings of the IEEE conference on computer vision and pattern recognition},
  pages={633--641},
  year={2017}
}

@article{codella2019skinisic,
  title={Skin lesion analysis toward melanoma detection 2018: A challenge hosted by the international skin imaging collaboration (isic)},
  author={Codella, Noel and Rotemberg, Veronica and Tschandl, Philipp and Celebi, M Emre and Dusza, Stephen and Gutman, David and Helba, Brian and Kalloo, Aadi and Liopyris, Konstantinos and Marchetti, Michael and others},
  journal={arXiv preprint arXiv:1902.03368},
  year={2019}
}

@inproceedings{demir2018deepglobe,
  title={Deepglobe 2018: A challenge to parse the earth through satellite images},
  author={Demir, Ilke and Koperski, Krzysztof and Lindenbaum, David and Pang, Guan and Huang, Jing and Basu, Saikat and Hughes, Forest and Tuia, Devis and Raskar, Ramesh},
  booktitle={Proceedings of the IEEE conference on computer vision and pattern recognition workshops},
  pages={172--181},
  year={2018}
}

@article{candemir2013lung1,
  title={Lung segmentation in chest radiographs using anatomical atlases with nonrigid registration},
  author={Candemir, Sema and Jaeger, Stefan and Palaniappan, Kannappan and Musco, Jonathan P and Singh, Rahul K and Xue, Zhiyun and Karargyris, Alexandros and Antani, Sameer and Thoma, George and McDonald, Clement J},
  journal={IEEE transactions on medical imaging},
  volume={33},
  number={2},
  pages={577--590},
  year={2013},
  publisher={IEEE}
}

@inproceedings{li2020fss,
  title={Fss-1000: A 1000-class dataset for few-shot segmentation},
  author={Li, Xiang and Wei, Tianhan and Chen, Yau Pun and Tai, Yu-Wing and Tang, Chi-Keung},
  booktitle={Proceedings of the IEEE/CVF conference on computer vision and pattern recognition},
  pages={2869--2878},
  year={2020}
}

@inproceedings{YueLZHLNDZJCJ24,
  author       = {Pengfei Yue and
                  Jianghang Lin and
                  Shengchuan Zhang and
                  Jie Hu and
                  Yilin Lu and
                  Hongwei Niu and
                  Haixin Ding and
                  Yan Zhang and
                  Guannan Jiang and
                  Liujuan Cao and
                  Rongrong Ji},
  editor       = {Jianfei Cai and
                  Mohan S. Kankanhalli and
                  Balakrishnan Prabhakaran and
                  Susanne Boll and
                  Ramanathan Subramanian and
                  Liang Zheng and
                  Vivek K. Singh and
                  Pablo C{\'{e}}sar and
                  Lexing Xie and
                  Dong Xu},
  title        = {Adaptive Selection based Referring Image Segmentation},
  booktitle    = {Proceedings of the 32nd {ACM} International Conference on Multimedia,
                  {MM} 2024, Melbourne, VIC, Australia, 28 October 2024 - 1 November
                  2024},
  pages        = {1101--1110},
  publisher    = {{ACM}},
  year         = {2024},
  url          = {https://doi.org/10.1145/3664647.3680850},
  doi          = {10.1145/3664647.3680850},
  bibsource    = {dblp computer science bibliography, https://dblp.org}
}

@inproceedings{ShiSS0YY24,
  author       = {Junjie Shi and
                  Caozhi Shang and
                  Zhaobin Sun and
                  Li Yu and
                  Xin Yang and
                  Zengqiang Yan},
  editor       = {Jianfei Cai and
                  Mohan S. Kankanhalli and
                  Balakrishnan Prabhakaran and
                  Susanne Boll and
                  Ramanathan Subramanian and
                  Liang Zheng and
                  Vivek K. Singh and
                  Pablo C{\'{e}}sar and
                  Lexing Xie and
                  Dong Xu},
  title        = {{PASSION:} Towards Effective Incomplete Multi-Modal Medical Image
                  Segmentation with Imbalanced Missing Rates},
  booktitle    = {Proceedings of the 32nd {ACM} International Conference on Multimedia,
                  {MM} 2024, Melbourne, VIC, Australia, 28 October 2024 - 1 November
                  2024},
  pages        = {456--465},
  publisher    = {{ACM}},
  year         = {2024},
  url          = {https://doi.org/10.1145/3664647.3681543},
  doi          = {10.1145/3664647.3681543},
  bibsource    = {dblp computer science bibliography, https://dblp.org}
}

@inproceedings{CaiSY24,
  author       = {Yuxiang Cai and
                  Yongheng Shang and
                  Jianwei Yin},
  editor       = {Jianfei Cai and
                  Mohan S. Kankanhalli and
                  Balakrishnan Prabhakaran and
                  Susanne Boll and
                  Ramanathan Subramanian and
                  Liang Zheng and
                  Vivek K. Singh and
                  Pablo C{\'{e}}sar and
                  Lexing Xie and
                  Dong Xu},
  title        = {MultiDAN: Unsupervised, Multistage, Multisource and Multitarget Domain
                  Adaptation for Semantic Segmentation of Remote Sensing Images},
  booktitle    = {Proceedings of the 32nd {ACM} International Conference on Multimedia,
                  {MM} 2024, Melbourne, VIC, Australia, 28 October 2024 - 1 November
                  2024},
  pages        = {1168--1177},
  publisher    = {{ACM}},
  year         = {2024},
  url          = {https://doi.org/10.1145/3664647.3681494},
  doi          = {10.1145/3664647.3681494},
  bibsource    = {dblp computer science bibliography, https://dblp.org}
}

@inproceedings{WuLLYZZ24,
  author       = {Yuanchen Wu and
                  Xiaoqiang Li and
                  Jide Li and
                  Kequan Yang and
                  Pinpin Zhu and
                  Shaohua Zhang},
  editor       = {Jianfei Cai and
                  Mohan S. Kankanhalli and
                  Balakrishnan Prabhakaran and
                  Susanne Boll and
                  Ramanathan Subramanian and
                  Liang Zheng and
                  Vivek K. Singh and
                  Pablo C{\'{e}}sar and
                  Lexing Xie and
                  Dong Xu},
  title        = {{DINO} is Also a Semantic Guider: Exploiting Class-aware Affinity
                  for Weakly Supervised Semantic Segmentation},
  booktitle    = {Proceedings of the 32nd {ACM} International Conference on Multimedia,
                  {MM} 2024, Melbourne, VIC, Australia, 28 October 2024 - 1 November
                  2024},
  pages        = {1389--1397},
  publisher    = {{ACM}},
  year         = {2024},
  url          = {https://doi.org/10.1145/3664647.3681710},
  doi          = {10.1145/3664647.3681710},
  bibsource    = {dblp computer science bibliography, https://dblp.org}
}

@inproceedings{SunZ024,
  author       = {Peiwen Sun and
                  Honggang Zhang and
                  Di Hu},
  editor       = {Jianfei Cai and
                  Mohan S. Kankanhalli and
                  Balakrishnan Prabhakaran and
                  Susanne Boll and
                  Ramanathan Subramanian and
                  Liang Zheng and
                  Vivek K. Singh and
                  Pablo C{\'{e}}sar and
                  Lexing Xie and
                  Dong Xu},
  title        = {Unveiling and Mitigating Bias in Audio Visual Segmentation},
  booktitle    = {Proceedings of the 32nd {ACM} International Conference on Multimedia,
                  {MM} 2024, Melbourne, VIC, Australia, 28 October 2024 - 1 November
                  2024},
  pages        = {7259--7268},
  publisher    = {{ACM}},
  year         = {2024},
  url          = {https://doi.org/10.1145/3664647.3680802},
  doi          = {10.1145/3664647.3680802},
  bibsource    = {dblp computer science bibliography, https://dblp.org}
}

@article{carion2025sam3,
  title={Sam 3: Segment anything with concepts},
  author={Carion, Nicolas and Gustafson, Laura and Hu, Yuan-Ting and Debnath, Shoubhik and Hu, Ronghang and Suris, Didac and Ryali, Chaitanya and Alwala, Kalyan Vasudev and Khedr, Haitham and Huang, Andrew and others},
  journal={arXiv preprint arXiv:2511.16719},
  year={2025}
}

@inproceedings{liu2025TVGT,
  title={Textual and visual guided task adaptation for source-free cross-domain few-shot segmentation},
  author={Liu, Jianming and Qiu, Wenlong and Wei, Haitao},
  booktitle={Proceedings of the 33rd ACM International Conference on Multimedia},
  pages={5150--5159},
  year={2025}
}

@InProceedings{Fan_2025_ICCV_isa,
    author    = {Fan, Qi and Liu, Kaiqi and Liu, Nian and Cholakkal, Hisham and Anwer, Rao Muhammad and Li, Wenbin and Gao, Yang},
    title     = {Adapting In-Domain Few-Shot Segmentation to New Domains without Source Domain Retraining},
    booktitle = {Proceedings of the IEEE/CVF International Conference on Computer Vision (ICCV)},
    month     = {October},
    year      = {2025},
    pages     = {21429-21439}
}

@inproceedings{siam2019amp,
  title={Amp: Adaptive masked proxies for few-shot segmentation},
  author={Siam, Mennatullah and Oreshkin, Boris N and Jagersand, Martin},
  booktitle={Proceedings of the IEEE/CVF International Conference on Computer Vision},
  pages={5249--5258},
  year={2019}
}

@inproceedings{zhang2019canet,
  title={Canet: Class-agnostic segmentation networks with iterative refinement and attentive few-shot learning},
  author={Zhang, Chi and Lin, Guosheng and Liu, Fayao and Yao, Rui and Shen, Chunhua},
  booktitle={Proceedings of the IEEE/CVF conference on computer vision and pattern recognition},
  pages={5217--5226},
  year={2019}
}

@inproceedings{yang2020rpmms,
  title={Prototype mixture models for few-shot semantic segmentation},
  author={Yang, Boyu and Liu, Chang and Li, Bohao and Jiao, Jianbin and Ye, Qixiang},
  booktitle={European conference on computer vision},
  pages={763--778},
  year={2020},
  organization={Springer}
}

@article{tian2020pfenet,
  title={Prior guided feature enrichment network for few-shot segmentation},
  author={Tian, Zhuotao and Zhao, Hengshuang and Shu, Michelle and Yang, Zhicheng and Li, Ruiyu and Jia, Jiaya},
  journal={IEEE transactions on pattern analysis and machine intelligence},
  volume={44},
  number={2},
  pages={1050--1065},
  year={2020},
  publisher={IEEE}
}

@article{zhang2023persam,
  title={Personalize Segment Anything Model with One Shot},
  author={Zhang, Renrui and Jiang, Zhengkai and Guo, Ziyu and Yan, Shilin and Pan, Junting and Dong, Hao and Gao, Peng and Li, Hongsheng},
  journal={arXiv preprint arXiv:2305.03048},
  year={2023}
}

@inproceedings{
tong2025adapterDFN,
title={Adapter Naturally Serves as Decoupler for Cross-Domain Few-Shot Semantic Segmentation},
author={Jintao Tong and Ran Ma and Yixiong Zou and Guangyao Chen and Yuhua Li and Ruixuan Li},
booktitle={Forty-second International Conference on Machine Learning},
year={2025},
}
%




\end{document}